\documentclass[journal]{IEEEtran}
\usepackage{framed,multirow}
\usepackage{graphicx}
\usepackage{wrapfig}
\usepackage{multirow}
\usepackage{epstopdf}
\usepackage{tabulary}
\usepackage{hyperref}
\usepackage{mathptmx} % assumes new font selection scheme installed
\usepackage{times} % assumes new font selection scheme installed
\usepackage{amsmath} % assumes amsmath package installed
\usepackage{amsfonts}
\usepackage{amssymb}  % assumes amsmath package installed
\usepackage{color}
\usepackage[bottom]{footmisc} % put footnote under figures
\usepackage{flushend} % aligned references columns

\definecolor{Gray}{gray}{0.25}
\definecolor{red}{rgb}{1.00,0.00,0.00}
\definecolor{blue}{rgb}{0.00,0.00,1.00}
\definecolor{green}{rgb}{0.0,0.5,0.1}
\definecolor{yellow}{rgb}{0.3,0.3,0.0}
\definecolor{pink}{rgb}{0.5,0.1,0.5}
\newcommand{\cred}[1] {\textcolor{red}{#1}}
\newcommand{\cblue}[1] {\textcolor{blue}{#1}}

\ifCLASSINFOpdf
\else
\fi

\begin{document}
\title{{\small \vspace{-1cm} \cblue{Accepted in IEEE/ASME Transactions on Mechatronics (TMECH) -- DOI: 10.1109/TMECH.2020.3048433 }}\\ \vspace{0.5cm} OrthographicNet: A Deep Transfer Learning Approach for 3D Object Recognition in Open-Ended Domains}

\author{S. Hamidreza Kasaei
	\thanks{Department of Artificial Intelligence, University of Groningen, Groningen, the Netherlands. Email: hamidreza.kasaei@rug.nl}
	\thanks{We thank NVIDIA Corporation for their generous donation of GPUs used for this research.}
		%\thanks{Manuscript received Jan. 21, 2020; %revised August 26, 2015.
	}

\maketitle

\begin{abstract}
	Nowadays, service robots are appearing more and more in our daily life. For this type of robot, open-ended object category learning and recognition is necessary since no matter how extensive the training data used for batch learning, the robot might be faced with a \emph{new object} when operating in a real-world environment. In this work, we present \emph{OrthographicNet}, a Convolutional Neural Network (CNN)-based model, for 3D object recognition in open-ended domains. In particular, \emph{OrthographicNet} generates a global rotation- and scale-invariant representation for a given 3D object, enabling robots to recognize the same or similar objects seen from different perspectives. Experimental results show that our approach yields significant improvements over the previous state-of-the-art approaches concerning object recognition performance and scalability in open-ended scenarios. Moreover, \emph{OrthographicNet} demonstrates the capability of learning new categories from very few examples on-site. Regarding real-time performance, three real-world demonstrations validate the promising performance of the proposed architecture. 
\end{abstract}


\section{Introduction}
\IEEEPARstart{H}umans can adapt to different environments dynamically by watching and learning about new object categories. In contrast, for migrating a robot to a new environment, one must often completely re-program the knowledge base that it is running with. 
Although many important problems have already been understood and solved successfully, many issues still remain. \emph{Open-ended object category learning} is one of these issues waiting for many improvements. 
In this work, open-ended learning refers to the ability to learn new object categories sequentially without forgetting the previously learned categories. In particular, the robot does not know in advance which object categories it will have to learn, which observations will be available, and when they will be available to support the learning. In such scenarios, the robot should learn about new categories from on-site experiences, supported in the feedback from human teachers. To address this problem, we believe the \emph{learning system} of the robots should have four characteristics~\cite{lopes2007many}: (\emph{i}) \textbf{on-line}: meaning that the learning procedure takes place while the robot is running; (\emph{ii}) \textbf{supervised}: to include the human instructor in the learning process. This is an effective way for a robot to obtain knowledge from a human teacher. (\emph{iii}) \textbf{incremental}: it is able to adjust the learned model of a certain category when a new instance is taught; (\emph{iv}) \textbf{opportunistic}: apart from learning from a batch of labeled training data at predefined times or according to a predefined training schedule, the robot must be prepared to accept a new example when it becomes available. This way, it is expected that the competence of the robot increases over time.

Most of the recent approaches use Convolutional Neural Networks (CNN) for 3D object recognition~\cite{tang2020onboard, kanezaki2018rotationnet, garcia2016pointnet, zhi2017lightnet, wu20153d,shi2015deeppano}. It is now clear that if an application has a pre-defined fixed set of object categories and thousands of examples per category, an effective way to build an object recognition system is to train a deep CNN. However, there are several limitations to use CNN in open-ended domains. CNNs are incremental by nature but not open-ended, since the inclusion of new categories enforces a restructuring in the topology of the network. 
\begin{figure}[t]
	\centering
	\includegraphics[width=\columnwidth, trim= 0cm 0cm 0cm 0cm,clip=true]{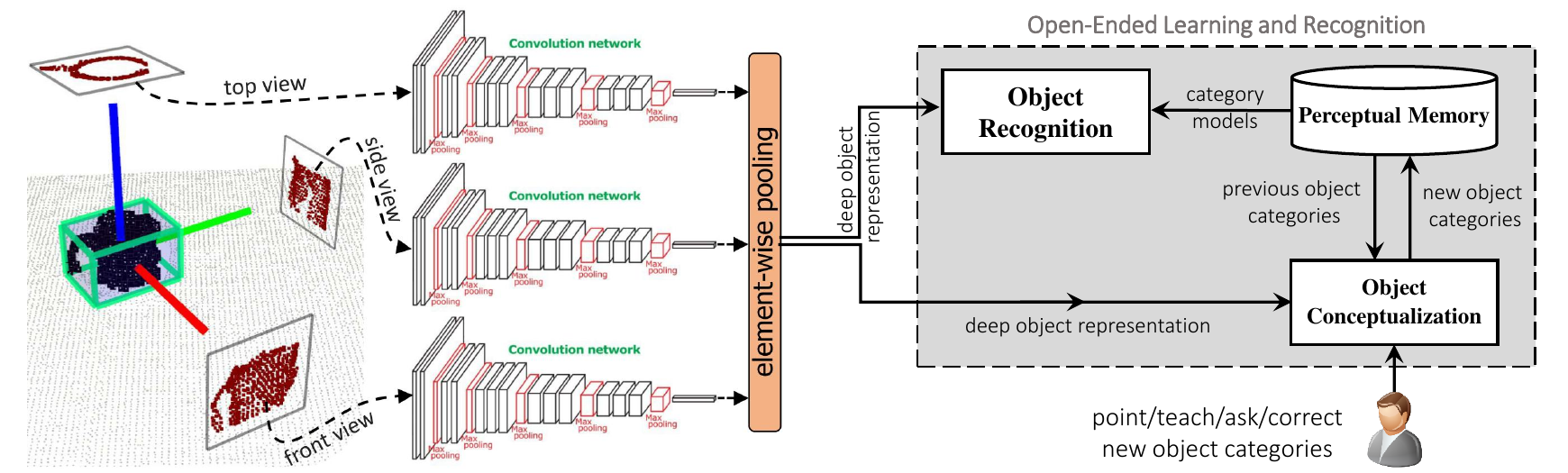}
	\caption{{Overview of the proposed open-ended object recognition pipeline: (\emph{left}) the
		\emph{cup} object and its bounding box, reference frame and three projected views; (\emph{center}) each orthographic projection is then fed to a CNN to obtain a view-wise deep feature; (\emph{right}) to construct a global deep representation for the given object, view-wise features are merged using an element-wise pooling function. The obtained representation is finally used for open-ended object category learning and recognition.}
		%\vspace{-3mm}
		}
	\label{fig:overal}
\end{figure}
Moreover, CNNs usually need a lot of training data, and if limited data is used for training, this might lead to non-discriminative object representations and, as a consequence, to poor object recognition performance. Deep transfer learning can relax these limitations and motivates us to combine deep-learned features with an online classifier to handle the problem of open-ended object category learning and recognition. To the best of our knowledge, there is no other deep learning based approach jointly tackling 3D object pose estimation and recognition in an open-ended fashion. 
As depicted in Fig.~\ref{fig:overal}, we first construct a unique reference frame for the given object. Afterwards, three principal orthographic projections including \emph{front}, \emph{top}, and \emph{right--side} views are computed by exploiting the object reference frame. Each projected view is then fed to a CNN to obtain a view-wise deep feature. To construct a global feature for the given object, the obtained view-wise features are then merged using an element-wise pooling function. The obtain feature is scale and pose invariant, informative and stable, and designed with the objective of supporting accurate 3D object recognition. We finally conducted our experiments with an instance-based learning. We perform a set of extensive experiments to assess the performance of OrthographicNet and compare it with a set of state-of-the-art methods.

The remainder of this paper is organized as follows: Section~\ref{sec:related_work} reviews related work of open-ended learning and deep learning approaches applied to 3D object recognition. Next, the detailed methodologies of our proposal -- namely \emph{OrthographicNet} -- are explained in Section~\ref{sec:object_representation} and \ref{sec:object_recognition}. Experimental result and discussion are given in Section~\ref{sec:results}, followed by conclusions in Section~\ref{sec:conclusion}.

%%%%%%%%%%%%%%%%%%%%%%%%%%%%%%%%%%%%%%%%%%%%%%%%%%%%%%%%%%%%%%%%%%%%%%%%%%%%%%%%%%%%%%%%%%%%%%%%%%%%%%
\section{Related work}
\label{sec:related_work}
%%%%%%%%%%%%%%%%%%%%%%%%%%%%%%%%%%%%%%%%%%%%%%%%%%%%%%%%%%%%%%%%%%%%%%%%%%%%%%%%%%%%%%%%%%%%%%%%%%%%%%

In the last decade, various research groups have made substantial progress towards the development of learning approaches which support online and incremental object category learning~\cite{kasaei2018perceiving,kasaei2018towards,faulhammer2017autonomous,kasaei2016hierarchical,oliveira20163d}. In such systems, \emph{object representation} plays a prominent role since its output uses in both learning and recognition phases. In \cite{kasaei2018perceiving}, an open-ended object category learning system, based on a global 3D shape feature namely GOOD~\cite{kasaei2016good}, is described. In particular, the authors proposed a cognitive robotic architecture to create a concurrent 3D object category learning and recognition in an interactive and open-ended manner. Kasaei et al.~\cite{kasaei2018towards} proposed a naive Bayes learning approach with a Bag-of-Words object representation to acquire and refine object category models in open-ended fashion. Faulhammer et al.~\cite{faulhammer2017autonomous} introduced a system which allows a mobile robot to autonomously detect, model and recognize objects in everyday environments. 
Skocaj et al.~\cite{skovcaj2016integrated} presented an integrated robotic system capable of interactive learning in dialogue with a human. Their system learns and refines conceptual models of visual objects and their properties, either by attending to information provided by a human tutor or by taking initiative itself. 
Oliveira et al.~\cite{oliveira2015concurrent} tackle this problem by proposing an approach for concurrent learning of visual code-books and object categories in open-ended domains. In \cite{kasaei2016hierarchical}, an extension of Latent Dirichlet Allocation (LDA) \cite{blei2003latent} namely Local-LDA is proposed that goes one step further by learning specific topics for each category independently and in an open-ended manner. All the above approaches use hand-crafted features. This in turn means that they may not generalize well across different domains. Recently, deep learning approaches have received significant attention from the robotics, machine learning, and computer vision communities. For 3D object recognition, deep learning approaches can be categorized into three categories according to their input: (\emph{i}) volume-based~\cite{wu20153d,maturana2015voxnet}, (\emph{ii}) view-based \cite{shi2015deeppano,sinha2016deep,su2015multi}, and (\emph{iii}) pointset-based methods~\cite{qi2017pointnet++,klokov2017escape}. Volume-based approaches, first represent an object as a 3D voxel grid and then use the obtained representation as the input to a CNN with 3D filter banks. Approaches of the second category (i.e., view-based) extract 2D images from the 3D representation by projecting the object's points into 2D planes. In contrast, pointset-based approaches work directly on 3D point clouds and require neither voxelization nor 2D image projections. Among these methods, view-based methods are shown effective in object recognition tasks and obtained the best recognition results so far~\cite{kanezaki2018rotationnet,Qi2016CVPR}. 
Our approach falls into the view-based category. Shi et al.~\cite{shi2015deeppano} tackle the problem of 3D object recognition by combining a panoramic representation of 3D objects with a CNN, i.e., named DeepPano. In this approach, each object is first converted into a panoramic view, i.e., a cylinder projection object around its principle axis. Then, a variant of CNN is used to learn the deep representations from the panoramic views. Su et al.~\cite{su2015multi} described an object recognition approach using multiple view-wise CNN features. In another work,  Sinha et al.~\cite{sinha2016deep} adopted an approach of converting the 3D object into a ``geometry image'' and used standard CNNs directly to learn 3D shape surfaces. These works are similar to ours in that they use multiple views of an object. However, unlike our approach, these approaches are not suitable for real-world scenarios since objects are often partially observable due to occlusions, which makes it difficult to rely on multi-view representations that are learned with the whole circumference. Several deep transfer learning approaches assumed that large amounts of training data are available for novel classes \cite{oquab2014learning2}. For such situations the strength of pre-trained CNNs for extracting features is well known \cite{oquab2014learning2,sharif2014cnn}.  Unlike our approach, CNN-based approaches are not scale and rotation invariant. Several researchers try to solve the issue by data augmentation either using Generative Adversarial Networks (GAN)~\cite{goodfellow2014generative} or by modifying images by translation, flipping, rotating and adding noise~\cite{wang2018low} i.e., CNNs are still required to learn the rotation equivariance properties from the data \cite{cohen2016group}. Unlike the above mentioned CNN-based approaches, we assume that the training instances are extracted from on-site experiences of a robot, and thus become gradually available over time, rather than being completely or partially available at the beginning of the learning process. Moreover, in our approach the set of classes is continuously growing while in the mentioned deep learning approaches the set of classes is predefined. Catastrophic forgetting is another important limitation of CNNs for being used in open-ended scenarios~\cite{kemker2018measuring}. In simple terms, the catastrophic forgetting problem is defined as forgetting the older categories as the robot is being trained further on newer categories. This usually happens due to modifying previously learned weights upon the start of training the network for newer categories. Several approaches have been proposed to tackle this problem; notable recent works include~{\cite{snell2017prototypical,wu2018memory,liu2019l3doc}}. Prototypical Networks learn to classify test examples by computing distances to prototype feature vectors of the novel categories~\cite{snell2017prototypical}. It should be noted that our approach is suitable for both instance-level (e.g., \textit{Mug1} vs. \textit{Mug2}) and category-level (e.g., \textit{Mug} vs. \textit{Plate}) object recognition, while Prototypical Networks can only be used for category-level object recognition. Unlike our approach,  \textit{Memory Replay} based approaches re-train the classifier once a new category is introduced, mainly using a large set of training instances of old categories and few training instances of the new category~\cite{snell2017prototypical}. {Lie et al.,~\cite{liu2019l3doc} proposed a deep transfer learning approach for lifelong 3D object classification (L3DOC). Similar to our approach, L3DOC can consecutively learn about new object categories mainly based on PointNet~\cite{pointNet2017}. The main idea of L3DOC is to extract and store the shared information of a set of known tasks and transfer the learned knowledge to the new coming task using a memory attention mechanism.} In general, such approaches are slow and not applicable in real-time robotic applications.

%%%%%%%%%%%%%%%%%%%%%%%%%%%%%%%%%%%%%%%%%%%%%%%%%%%%%%%%%%%%%%%%%%%%%%%%%%%%%%%%%%%%%%%%%%%%%%%%%%%%%%
%%%%%%%%%%%%%%%%%%%%%%%%%%%%%%%%%%%%%%%%%%%%%%%%%%%%%%%%%%%%%%%%%%%%%%%%%%%%%%%%%%%%%%%%%%%%%%%%%%%%%%
\section{Object representation}
\label{sec:object_representation}
%%%%%%%%%%%%%%%%%%%%%%%%%%%%%%%%%%%%%%%%%%%%%%%%%%%%%%%%%%%%%%%%%%%%%%%%%%%%%%%%%%%%%%%%%%%%%%%%%%%%%%

\begin{figure}[!b]
	\centering
	\includegraphics[width=0.95\linewidth, trim= 0cm 0cm 0cm 0cm,clip=true]{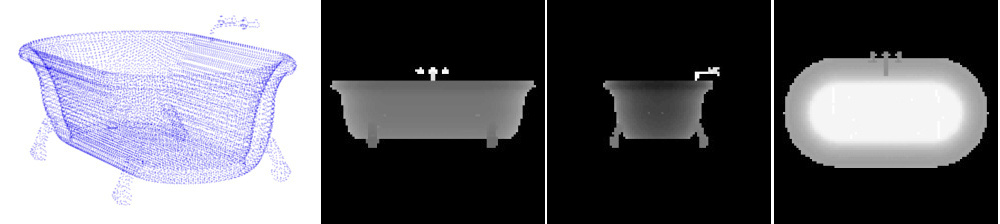}
	\caption{A sample point cloud of a bathtub and its orthographic projections.}
	\label{object}
\end{figure}
A point cloud of an object is represented as a set of points, $\textbf{p}_i : i \in \{1, \dots, n\}$, where each point is described by their 3D coordinates [x, y, z] and [R, G, B] color information. As shown in Fig.~\ref{fig:overal} (\emph{left}), OrthographicNet starts with constructing a \emph{global object reference frame} (RF) for a given object. Towards this end, three principal axes of the object are constructed based on eigenvectors analysis. In particular, we first compute the geometric center of the object using $\textbf{c} = \frac{1}{n}\sum_{i=1}^{n} \textbf{p}_i$. The normalized covariance matrix, $\Sigma$, of the object is then calculated by $\Sigma = \frac{1}{n}\sum_{i=1}^{n} (\textbf{p}_i-\textbf{c})(\textbf{p}_i-\textbf{c})^T$. Afterwards, eigenvalue decomposition is performed by $\Sigma \bold{V} = \bold{E}\bold{V}$, where $\bold{V} =[\vec{v_1},\vec{v_2},\vec{v_3}]$ is eigenvectors of $\Sigma$, and $\bold{E}= [\lambda_1,\lambda_2,\lambda_3]$ is the corresponding eigenvalues and $\lambda_1\ge\lambda_2\ge\lambda_3$. 
In this case, the largest eigenvector, $\vec{v_1}$, of the covariance matrix always points into the direction of the largest variance of the object's points, and the magnitude of this vector equals the corresponding eigenvalue, $\lambda_1$. The second largest eigenvector, $\vec{v_2}$, is always orthogonal to the largest eigenvector, and points into the direction of the second largest spread of the data. Therefore, the first two axes, X and Y, are defined by the eigenvectors $\vec{v_1}$ and $\vec{v_2}$, respectively. However, regarding the third axis, Z, instead of defining
it based on $\vec{v_3}$, we define it based on the cross product $\vec{v_1} \times \vec{v_2}$. The object is then transformed to be placed in the reference frame. 

Afterwards, we use an orthographic projection method to generate views of the object. It is worth to mention that the orthographic projection is a universal language among people in engineering professions and uses for technical drawing. In orthographic projection, up to six views of an object can be produced (called primary views). In this work, we just use three projection views including \emph{front}, \emph{top}, \emph{right-side} and do not consider the \emph{rear}, \emph{bottom} and \emph{left-side} views since they are mirror of the considered views. In this method, since projection lines are parallel to each other and are perpendicular to the projection plane, an accurate outline of the visible face of the object is obtained. We therefore create three square projection planes centered on the object's center. Each plane of projection is positioned between the observer and the object and is perpendicular to one axis and parallel with the others axes of the object reference frame. The side length of projection planes, $l$, is determined by the largest edge length of a tight-fitting axis-aligned bounding box (AABB) of the object. This choice makes the projections scale invariant. 
The dimensions of the AABB are obtained by computing the minimum and maximum coordinate values along each axis. The object is then projected on the planes. As an example, three orthographic projections of a bathtub object are shown in Fig.~\ref{object}. Afterwards, the obtained projection views are uniformly scaled to an appropriate input image size and then fed into a CNN to extract view-wise features. 

\begin{figure}[t]
	\centering
	\includegraphics[width=\columnwidth, trim= 0cm 0.0cm 1cm 0cm,clip=true]{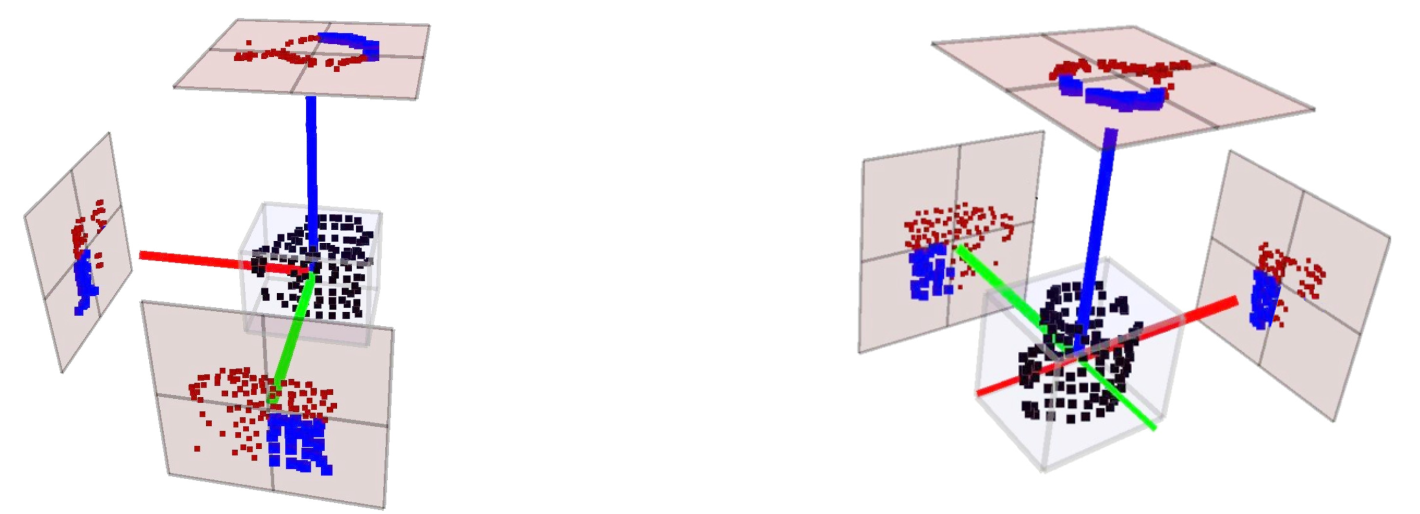}
	\caption{Visualization of sign disambiguation procedure: The red, green and blue lines represent the unambiguous X, Y, Z axes respectively. Two orthographic projections of the object on the XoZ and YoZ planes are used to determine the sign of X and Y axis respectively; (\emph{left}) sign is positive and; (\emph{right}) sign is negative, and therefore all projections have been mirrored. For a better representation, we highlighted the corresponding bins.}
	\label{fig:sign}
\end{figure}

It should be noted that the direction of eigenvectors is not unique, i.e., not repeatable across different trials  and its orientation has $180^\circ$ ambiguity. Therefore, the orthographic projections can be mirrored as shown in Fig.~\ref{fig:sign}. This problem is known as the \emph{sign ambiguity}, for which there is no mathematical solution~\cite{bro2008resolving}. To cope with this issue, it is suggested that the sign of each axis be similar to the sign of the Pearson's correlation of the scatter points. For building a scatter plot, each projected point $\rho = (\alpha, \beta) \in R^2$, where $\alpha$ is the perpendicular distance to the horizontal axis and $\beta$ is the perpendicular distance to the vertical axis, is then shifted to right and top by $\frac{l}{2}$. To complete the disambiguation, Pearson's correlation, $r$, is computed for the XoZ projection plane to find the direction of X axis:

\begin{equation}
r_{x} = \frac{\sum \alpha_{i}\beta_{i}~-~n{\bar {\alpha}}{\bar {\beta}}}{\sqrt {(\sum \alpha_{i}^{2}-n{\bar {\alpha}}^{2})}~~{\sqrt {(\sum \beta_{i}^{2}-n{\bar {\beta}}^{2})}}}.
\end{equation}

\noindent
where $\bar{\alpha}$ and $\bar {\beta}$ are the mean of $\alpha$ and $\beta$ respectively. In particular, Pearson's correlation reflects the non-linearity and direction of a linear relationship as a value between $-1$ and $1$, where $1$ indicates a strong positive relationship, $-1$ indicates a strong negative relationship. A similar indication, $r_{y}$, is computed for the Y axis using YoZ plane. Finally, the sign of the axes is determined as $s = r_{x} . r_{y}$, where s can be either positive or negative. In the case of negative $s$, the three projections should be mirrored otherwise not. Therefore, the final RF (X, Y, Z) is defined by $(s\vec{v_1},~s\vec{v_2},~\vec{v_1} \times \vec{v_2})$.

%%%%%%%%%%%%%%%%%%%%%%%%%%%%%%%%%%%%%%%%%%%%%%%%%%%%%%%%%%%%%%%%%%%%%%%%%%%%%%%%%%%%%%%%%%%%%%%%%%%%%%
\section{Object category learning and recognition}
\label{sec:object_recognition}
%%%%%%%%%%%%%%%%%%%%%%%%%%%%%%%%%%%%%%%%%%%%%%%%%%%%%%%%%%%%%%%%%%%%%%%%%%%%%%%%%%%%%%%%%%%%%%%%%%%%%%

{Object recognition using very limited training data is crucial for many computer vision and robotics applications and has recently attracted interests in both research communities and industries. This problem becomes even harder when the system must learn about new object categories from very few examples online while maintaining the recognition performance on previously learned categories. As an example, consider a robotic scenario, in which a robot faced with a new object category. It would be ideal if a user can teach the new category to the robot on-site using a few object views from the new category without accessing to the old training data used for previously learned categories. It is very difficult to use CNNs for such tasks since they are data hungry approaches and suffering from catastrophic forgetting as we discussed in previous sections. We have tackled this problem by proposing an instance-based learning and recognition (IBL) approach which considers category learning as a process of learning about the instances of a category, i.e., a category is represented simply by a set of known instances, $\textbf{C} \leftarrow \{ \textbf{O}_1, \dots, \textbf{O}_n\}$, where $\textbf{O}_i$ is the deep representation of an object view. In this work, the deep representation of an object is obtained after the element-wise pooling layer as shown in Fig.~\ref{fig:overal}.}
IBL is a baseline approach to evaluate object representations. An advantage of the IBL approaches over other machine learning methods is the ability of fast adapting an object category model to a previously unseen instance by storing the new instance or by throwing away an old instance. Therefore, the training phase is very fast in IBL approaches. Such learning approaches are able to recognize objects using a small number of \emph{representative} instances, while storing too many \emph{redundant} instances results in large memory consumption and slows down the recognition speed. Therefore, in our current setup a new instance of a specific category is stored in the robot's memory in the following situations:

\begin{itemize}
	\item  When the teacher for the first time teaches a certain category, through a \emph{teach} or a
	\emph{correct} action, an instance-based model of this new category is created and initialized with the set of views of the target object collected since object tracking started:
	
	\begin{equation}
	\textbf{C}_1 \leftarrow \{ \textbf{O}_1, \dots, \textbf{O}_{1k_1}\},
	\end{equation}
	where $k_1$ is the number of stored key object views for the first teaching action.
	
	\item  In later teaching actions, the target object views are added to the model of the category:
	\begin{equation}
	\textbf{C}_n \leftarrow \textbf{C}_{n-1} \cup \{ \textbf{O}_{nk}, \dots, \textbf{O}_{1k_n}\},
	\end{equation}
	where $k_n$ is the number of stored key object views for the n-th teaching action.
	
\end{itemize}

\noindent
{
It is worth mentioning that it is possible to let the robot to learn in a self-supervise manner. For instance, if the dissimilarity of a new object to all known categories is above a threshold, the robot infers the given object is not belong to known categories, and thus initializes a new category labeled as ``\emph{category\_m+1}'', where $m$ is the number of known categories so far. However, such a category label is not meaningful for a human user, and the user is not able to naturally instruct the robot to perform a task, e.g., \emph{"give me a mug"}. In contrast, the mechanism of introducing a new object category through human-robot interaction not only makes a mutual language for robot and user, but also prevents linearly accumulation of redundant object views and optimizes the number of instances required to represent a specific category leading to less memory consumption\footnote{{It is worth mentioning that prototype learning approaches~\cite{snell2017prototypical} or forgetting mechanisms \cite{shmelkov2017incremental} can be used to bound the memory usage of each category and the overall memory usage.}}.} Whenever a new object is added to a category, the agent retrieves the current model of the category and updates it by adding the representation of new object views to the category model. In particular, our approach can be seen as a combination of a particular \emph{object representation}, \emph{similarity measure} and \emph{classification rule}. Therefore, the choice of the similarity metric has an impact on the recognition performance. 
In the case of similarity measure, since each object is represented by a global deep feature, the dissimilarity between two objects can be computed by different distance functions.

We refer the reader to a comprehensive survey on distance/similarity measures provided by S. Cha~\cite{cha2007comprehensive}. After performing several cross-validation experiments, we conclude two type of distance functions including Jensen-Shannon (JS) and chi-squared ($\chi^{2}$) distances are suitable to estimate the similarity between two instances. Both functions are in the form of a bin-to-bin distance function. Although the practical results of $\chi^{2}$ and JS are almost identical, $\chi^{2}$ is computationally more efficient. Therefore, we use $\chi^{2}$ function to estimate the similarity of two instances. Mathematically, let $\operatorname{P}$ and $\operatorname{Q}$ $\in{\rm I\!R^{K}}$ be the representation of two objects:

\begin{equation}
\operatorname{D}\operatorname{(P,Q)} = \frac{1}{2}\sum_{i=1}^{K} \frac{(\operatorname{P}_i-\operatorname{Q}_i)^2}{(\operatorname{P}_i+\operatorname{Q}_i)},\label{chi}
\end{equation}

The dissimilarity of a target object and all stored instances of all categories, $\textbf{C}_i$, is computed using Eq.~\ref{chi}. The target object is finally classified based on the nearest neighbor distance: 

\begin{equation}
\operatorname{catgeory}(\operatorname{\textbf{T}}) = \underset{\textbf{C}\in\{1, \dots, m\}}{argmin} ~ \operatorname{D}(\operatorname{\textbf{T}}, \operatorname{\textbf{O}}_{\operatorname{\textbf{C}}_i}).
\end{equation}

\noindent where $m$ is the number of known categories up to now, and $\operatorname{\textbf{O}}_{\operatorname{\textbf{C}}_i}$ is the $i^{th}$ stored instances of category $\textbf{C}$.

%%%%%%%%%%%%%%%%%%%%%%%%%%%%%%%%%%%%%%%%%%%%%%%%%%%%%%%%%%%%%%%%%%%%%%%%%%%%%%%%%%%%%%%%%%%%%%%%%%%%%%
\section{Experimental result and discussion}
\label{sec:results}
%%%%%%%%%%%%%%%%%%%%%%%%%%%%%%%%%%%%%%%%%%%%%%%%%%%%%%%%%%%%%%%%%%%%%%%%%%%%%%%%%%%%%%%%%%%%%%%%%%%%%%

We extensively evaluate the proposed recognition system using {four} types of experiments. {All tests were performed with a PC running Ubuntu 18.04 with a 3.20 GHz Intel Xeon(R) i7 CPU, and a Quadro P5000 NVIDIA graphics card.}

\subsection{Ablation Study}
As shown in Fig.~\ref{fig:overal}, the proposed approach uses orthographic projections and three CNNs to generate a global deep representation for the given object. In this approach, the resolution of orthographic projections and the CNN architecture must be well selected to provide a good balance among recognition performance, computation time and memory usage. To define the optimal system configuration, we conduct 14 sets of experiments using various CNN architectures. All CNNs have been pre-trained on ImageNet dataset~\cite{deng2009imagenet} using the default network configurations as suggested in their scientific papers. In this work,  we assumed that the system does not have predefined set of training data for the target task (3D object recognition) available at the beginning of the learning process. It is worth mentioning if an application has enough training data in advance, even random initialization could produce excellent results~\cite{he2018rethinking}. For each CNN, 18 experiments were performed with different resolutions of orthographic images (i.e., ranging from $25\times25$ to $225\times225$ pixels) and avg/max pooling functions. 

%%%%%%%%%%%%%%%%%%%%%%%%%%%%%%%%%%%%%%%%%%%%%%%%%%%
\textbf{Dataset and evaluation metrics: } 
%%%%%%%%%%%%%%%%%%%%%%%%%%%%%%%%%%%%%%%%%%%%%%%%%%%
The offline evaluations were carried out using \emph{Princeton ModelNet10} dataset~\cite{wu20153d}. This dataset categorized 4899 objects into ten categories and defined training and testings data for each category (i.e., in total 3991 training samples and 908 testing samples). ModelNet10 is a relatively small dataset with significant intra-class variation; therefore, it is suitable for performing extensive sets of experiments to fine-tune the parameters of the system. 
The obtained results is reported using two metrics: Average Instance Accuracy (AIA) and Average Class Accuracy (ACA). The former one is defined as the percentage of the correctly recognized test instances. The latter one, i.e., ACA, is used to represent the average accuracy of all categories. 

\begin{figure}[b]
	\center
	\begin{tabular}[width=\linewidth]{cc}
		\hspace{-5mm}
		\includegraphics[width=0.49\linewidth, trim= 0.2cm 0.65cm 0.2cm 0.5cm,clip=true]{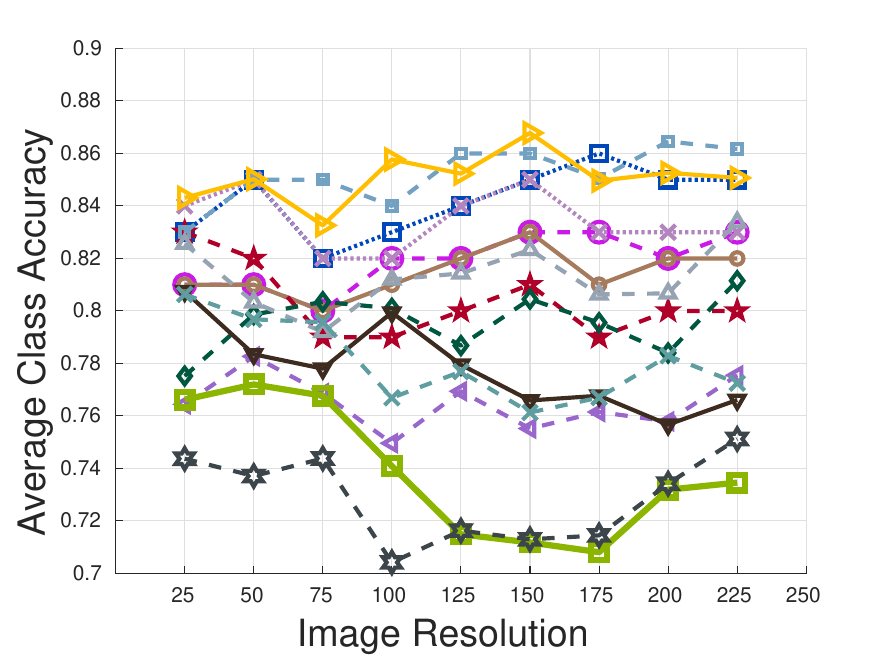} & \hspace{-8mm}
		\includegraphics[width=0.568\linewidth, trim= 1.5cm 0.65cm 0.8cm 0.7cm,clip=true]{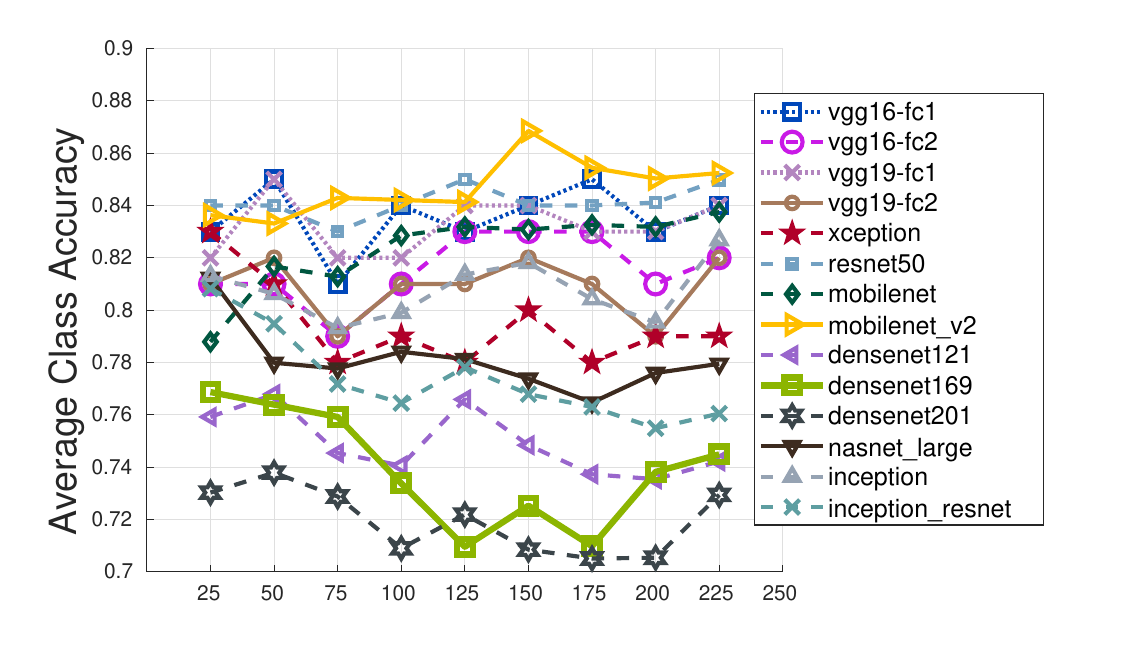}\\
		\hspace{-5mm} 
		\includegraphics[width=0.49\linewidth, trim= 0.2cm 0cm 0.2cm 0cm,clip=true]{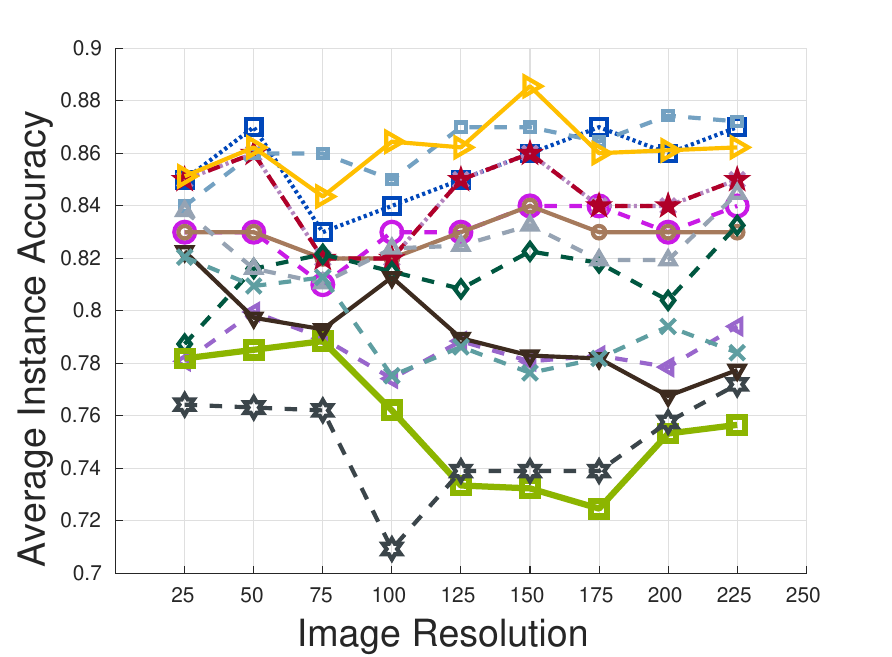} & \hspace{-9mm}
		\includegraphics[width=0.568\linewidth, trim= 1.65cm 0.03cm 1.45cm 0.5cm,clip=true]{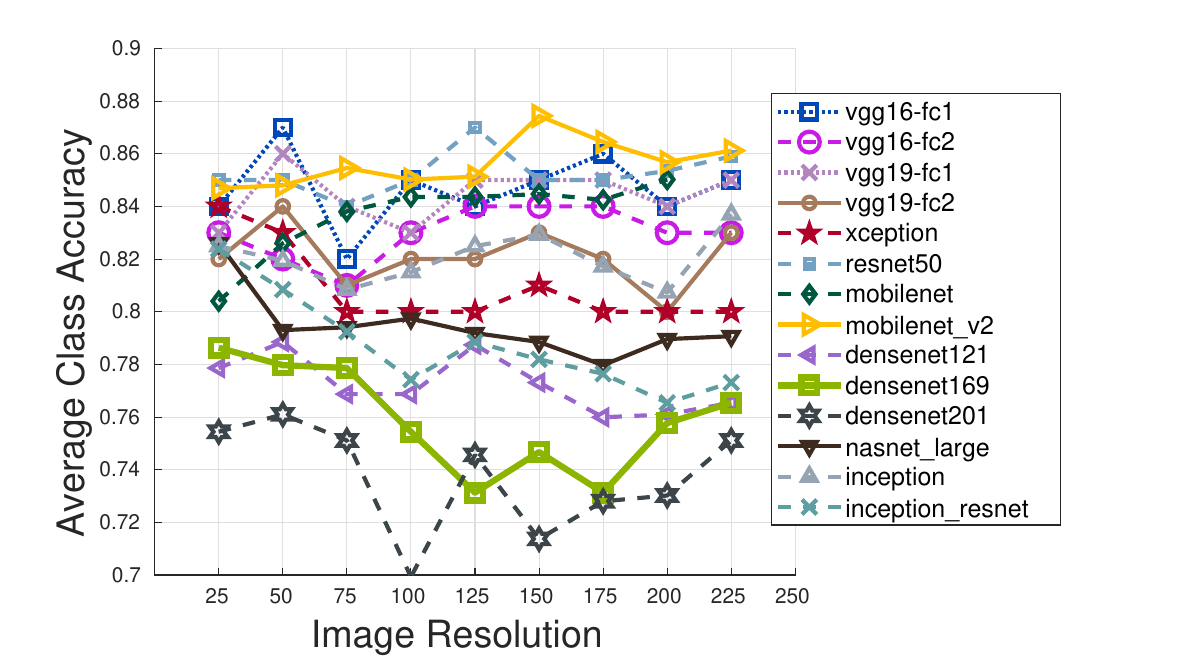}\\
	\end{tabular}
	\caption{Summary of off-line evaluations using various CNN architectures and two pooling functions: (\emph{top-row}) average class accuracy vs. image resolution; (\emph{bottom-row}) average instance accuracy vs. image resolution; (\emph{left-column}) results using max pooling; (\emph{right-column}) results using average pooling.}
	\label{fig:offline_evaluations}       % Give a unique label
\end{figure}

%%%%%%%%%%%%%%%%%%%%%%%%%%%%%%%%%%%%%%%%%%%%%%%%%%%
\textbf{Results: }
%%%%%%%%%%%%%%%%%%%%%%%%%%%%%%%%%%%%%%%%%%%%%%%%%%
A summary of the experiments is plotted in Fig.~\ref{fig:offline_evaluations}. In the case of ACA (\emph{top-row}), the best result was obtained with \emph{MobileNet-v2}, \emph{average pooling} and image resolution of $150\times150$ pixels. The accuracy of the proposed system with this configuration was 0.8685. It should be mentioned that while a higher resolution orthographic images may provide more details about the object, it increases computation time, memory usage and sensitivity to noise. {As it can be observed from Fig.~\ref{fig:offline_evaluations} (\emph{top-row}), the {average class accuracy} of \emph{DenseNet} architectures~\cite{huang2017DenseNet} (i.e., 121, 169, 201) is the worst among the evaluated CNNs and \emph{MobileNet-v2}, \emph{VGG16-fc1}~\cite{simonyan2014VGG} and \emph{ResNet50}~\cite{he2016ResNet} are the top-three CNNs, achieving a good average class accuracy with stable performance. An important observation is that, although \textit{DenseNet} architectures are more advanced models, their accuracies are not as good as the others. This observation can be explained by this fact that deeper networks learn a set of specific features (problem dependent), while less complex networks usually learn general features (problem independent). In other words, we observed that since orthographic images are different from images in ImageNet dataset (used to pre-train networks), general deep features learned by small networks could represent orthographic images much better than specific deep features learned by deep networks. We also realized that several misclassifications mainly occurred among items that look alike. In particular, some instances in the \emph{desk} category has a very similar shape to the instances of \emph{table} category; similarly, there are several highly similar instances in \emph{dresser} and \emph{night\_stand} categories.}

\begin{table}[!t]
	\begin{center}
		\caption {Proprieties of the top-three CNNs\vspace{-2mm}}
		\resizebox{\linewidth}{!}{
			\begin{tabular}{ |c|c|c|c|c|c| }
				\hline
				Model & Feature Length & Size & \#Parameters & Depth \\
				\hline \hline
				MobileNet-v2	& 1280 float 	&	14 MB	&	3.53 M		&	88\\\hline
				VGG16	    	& 4096 float	&	528 MB	& 	138.35 M	&	23\\\hline
				ResNet50   		& 4096 float	&	99 MB	&	25.63 M		&	50\\
				\hline 
		\end{tabular}}
		\label{table:CNNs}
	\end{center}
	%\vspace{-5mm}
\end{table}
Fig.~\ref{fig:offline_evaluations}~(\emph{bottom-row}) compares the average instance accuracy of the evaluated system configurations in which several observations can be made. First, similar to the previous round of experiments, our approach with \emph{MobileNet-v2}, \emph{VGG16-fc1} and \emph{ResNet50} demonstrated a good performance. In contrast, the AIA of DenseNets was lower than the others under all level levels of image resolution. The other evaluated CNNs, including \emph{Xception}~\cite{chollet2017Xception}, \emph{Nasnet}~\cite{zoph2018Nasnet}, \emph{Inception}~\cite{szegedy2016Inception} and \emph{Inception\_Resnet}~\cite{szegedy2017InceptionResnet}, demonstrated a medium-level performance. Overall, the best AIA obtained with \emph{MobileNet-v2}, \emph{max pooling} and orthographic projections with the resolution of $150\times150$ pixels which was 88.56 percent. These parameters were selected as the default system configurations. The AIA of the system with the default configuration and \emph{average pooling} was 87.44. Note that the length of feature vector, size and number of parameters of the network have direct influence on memory usage and computation time. {Table~\ref{table:CNNs} summarizes the proprieties of the top-three CNNs.} The length of feature extracted by \emph{MobileNet-v2} is $1280$ float, while both \emph{VGG16-fc1} and \emph{ResNet50} represent an object by a vector of $4096$ float. The size of \emph{MobileNet-v2} is 14MB which is around $37$ and $7$ times smaller than than \emph{VGG16} and \emph{ResNet50} respectively. {We also performed a experiment to measure the computational time of the proposed approach with MobileNet-V2. Since the number of object's points directly effects the computational time, we calculate the average time required to render three projections and process them by the network for $20$ randomly selected objects from the ModelNet10 dataset. The average computation time for rendering orthographic images was $0.003$ second, and processing all three views by the network to generate a global feature for the given object on average took $0.008$ second. According to these evaluations, our approach with \emph{MobileNet-v2} is competent for applications with strict limits on the memory and computation time requirements.} This evaluation shows that the overall performance of the proposed approach is promising and it is capable of providing distinctive global features for 3D objects. 

%%%%%%%%%%%%%%%%%%%%%%%%%%%%%%%%%%%%%%%%%%%%%%%%%%%%%%%%%%%
\subsection{Offline Evaluation}
%%%%%%%%%%%%%%%%%%%%%%%%%%%%%%%%%%%%%%%%%%%%%%%%%%%%%%%%%%%

{According to the results of our ablation study, we used the MobileNet-V2 as the backbone of OrthographicNet architecture. Furthermore, to have a fair comparison with the state-of-the-art approaches, we replaced the open-ended classifier of the proposed architecture with a small Deep Neural Network (DNN). We trained 100 networks with different architectures, various optimization algorithms, and different learning rate for 200 epochs each. The network architecture that obtained the best performance consists of four fully connected layers with $1024$, $512$, $256$, $128$ neurons respectively, and a softmax layer with the same number of nodes as the number of classes exist in the object dataset. We used a relu activation function for each fully connected layer. A dropout layer has been added between each pairs of fully connected layers, as well as the last fully connected layer and the softmax layer. We set the dropout rate to $30\%$ in all layers. In this round of experiments, OrthographicNet is trained from scratch using stochastic gradient descent (SGD) with a momentum of $0.9$. The learning rate, $\eta$, is initially set to $0.1$, and decayed by $\eta \times (0.98 \times \#\operatorname{epochs})$. Note that all weights are initialized randomly and the categorical cross-entropy is used as loss.}

%%%%%%%%%%%%%%%%%%%%%%%%%%%%%%%%%%%%%%%%%%%%%%%%%%%
\textbf{Dataset and evaluation metrics: } 
%%%%%%%%%%%%%%%%%%%%%%%%%%%%%%%%%%%%%%%%%%%%%%%%%%%
We evaluated our approach using ModelNet10 and ModelNet40 datasets. The the former dataset has been discussed in the previous subsection. The later dataset, ModelNet40, consists of $12311$ instances categorized in $40$ categories, in which there are $9843$ training and $2468$ test objects~\cite{wu20153d}. It should be noted that we used the original train and test split of datasets~\cite{wu20153d}, and generated three orthographic projects (top, side, and front views) of all training and testing objects. 
%\cred{We made the orthographic views of all objects of these datasets available online}\footnote{\cred{github link to the datasets}}.  
The classification accuracy is used as evaluation metric to compare the performance of different approaches.

%%%%%%%%%%%%%%%%%%%%%%%%%%%%%%%%%%%%%%%%%%%%%%%%%%%
\textbf{Results: } 
%%%%%%%%%%%%%%%%%%%%%%%%%%%%%%%%%%%%%%%%%%%%%%%%%%%
{We compared our method with 13 state-of-the-art approaches. Table~\ref{table:offline_STOA}
summarizes the comparative results of classification on ModelNet10 and ModelNet40. In the case of ModelNet10, 
it was observed that the classification accuracy of our approach was significantly better than the other state-of-the-art approaches that use 2D projection as input. Particularly, it was $1.47$ percentage points (p.p.) and $1.77$ p.p. better than RotationNet with three views and Pairwise, i.e., the two best approaches among 2D projections based approaches, respectively. Furthermore, \textit{OrthographicNet} significantly outperformed the state-of-the-art methods that use 3D volumetric data and point clouds as input. In particular, OrthographicNet clearly outperforms both VRN~\cite{brock2016generative} and LightNet~\cite{zhi2017lightnet} methods by $0.97$ p.p. and $1.18$ p.p., respectively. In the case of the ModelNet40 dataset, the proposed OrthographicNet method outperformed state-of-the-art approaches that use 2D projections and most of the approaches that use 3D data as input (outperformed $12$ out of $13$ approaches). However, it was observed that VRN slightly worked better than our approach (i.e., $91.00$ vs. $91.33$). The fact that our approach is computed on a stable, unique, and unambiguous local reference frame is likely to explain the obtained results. Furthermore, our approach represents a 3D object using three orthographic projections. In particular, OrthographicNet (\textit{i}) provides a compact, unique, and rich representation of objects, and (\textit{i}) is efficient to compute and less affected by noise since in each projection one of the dimensions of the object is discarded.}

\begin{table}[!t]
	\begin{center}
		\caption {Classification accuracy and comparison to state-of-the-art methods using ModelNet datasets. }
		\resizebox{\linewidth}{!}{
			\begin{tabular}{ |c|c|c|c|c|c| }
				\hline
				 Input data & Algorithm & ModelNet40 & ModelNet10 \\\hline
				 \multirow{8}{*}{2D projection} & MVCNN~\cite{su2015multi}  & 90.10 & -- \\\cline{2-4}
				 & Pairwise~\cite{johns2016pairwise}  & 90.70 & 92.80 \\\cline{2-4}
				 & PANORAMA-NN~\cite{sfikas2017exploiting} & 90.7 & 91.1 \\\cline{2-4}
				 & Geometry Image~\cite{sinha2016deep}  & 83.9 & 88.4 \\\cline{2-4}
				 & Multiple Depth Maps~\cite{zanuttigh2017deep} & 87.8 & 91.5 \\\cline{2-4}
				 & DeepPano~\cite{shi2015deeppano} & 77.63 & 85.45 \\\cline{2-4}
				 & GIFT~\cite{bai2016gift} &    83.10 & 92.35 \\\cline{2-4}
				 & RotationNet (3 Views)~\cite{kanezaki2018rotationnet} & 89.00 & 93.10 \\\cline{2-4}
				 & \textbf{OrthographciNet} &    \cblue{91.00} & \textbf{94.57} \\\cline{2-4}
				 \hline
				\multirow{4}{*}{3D Volume} & 3DShapeNets~\cite{wu20153d}  & 77.0 & 83.50 \\\cline{2-4}
				& 3D-GAN~\cite{wu2016learning} & 83.30 & 91.00 \\\cline{2-4}
				& LightNet~\cite{zhi2017lightnet} & 86.90 & 93.39 \\\cline{2-4}
				& FusionNet~\cite{hegde2016fusionnet} & 90.80 & 93.11 \\\cline{2-4}
			    & \textbf{VRN}~\cite{brock2016generative} & \textbf{91.33} & \cblue{93.60} \\\cline{2-4}
				\hline 
		\end{tabular}}
		\label{table:offline_STOA}
		\\\scriptsize{* The best result is shown in \textbf{bold}, and the second best is highlighted by \cblue{\textit{blue color}}. }
	\end{center}
\end{table}
%%%%%%%%%%%%%%%%%%%%%%%%%%%%%%%%%%%%%%%%%%%%%%%%%%
\subsection{Open-Ended Evaluation}
%%%%%%%%%%%%%%%%%%%%%%%%%%%%%%%%%%%%%%%%%%%%%%%%%%

Offline evaluation methodologies follow the classical \texttt{Train-then-Test} procedure, i.e., two separate stages. These methodologies are not well suited to evaluate open-ended learning systems, because they do not abide to the simultaneous nature of learning and recognition and also the number of categories must be predefined in such evaluations. Therefore, another round of experiments was carried using an open-ended teaching protocol~\cite{kasaei2015interactive,chauhan2011using,kasaei2018coping}. The main idea is to emulate the interactions of a robot with the surrounding environment over long periods of time. The teaching protocol determines which examples are used for training the algorithm, and which are used to test the algorithm. This protocol is based on a \texttt{Test-then-Train} scheme, which can be followed by a human user or by a simulated user. We developed a \emph{simulated teacher} to perform a set of consistent and reproducible experiments  in a fraction of time a human
would take to do the same task. The simulated user interacts with the agent using three basic actions:
\begin{itemize}
	\item \textbf{Teach}: introduces the category of an object to the agent,
	\item \textbf{Ask}: to inquire about the category of a test object view,
	\item \textbf{Correct}: if the agent classifies an object view incorrectly, the simulated teacher provides corrective feedback.
\end{itemize}

\begin{figure}[!t]
	\centering
	\includegraphics[width=\columnwidth, trim= 0cm 0cm 0cm 0cm,clip=true]{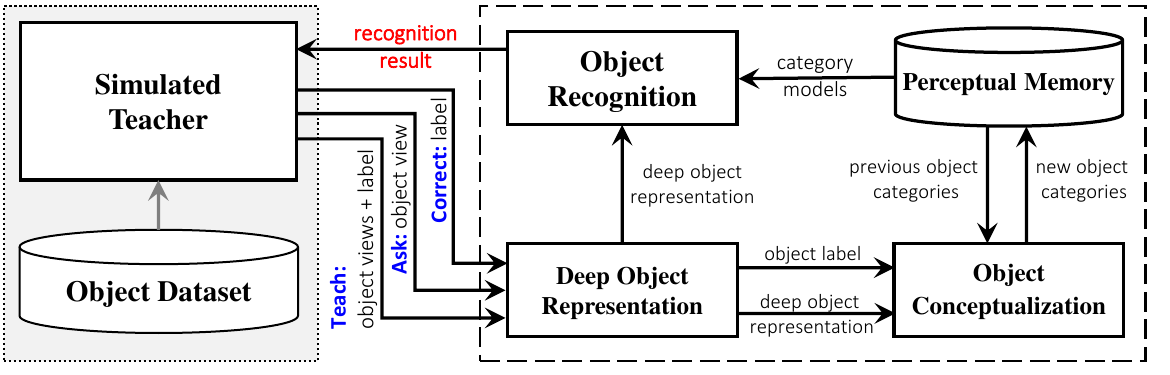}
	\caption{{Interaction between the simulated teacher and the learning agent; (\emph{left}) The simulated teacher is connected to a large object dataset and interacts with the agent by \emph{teach}, \emph{ask} and \emph{correct} actions as shown by blue color; (\emph{right}) In case of ask action, the agent is evaluated by a never-seen-before object. The agent recognizes the object and sends back the result to the simulated user (shown by red color); In the cases of teach and correct actions, the agent creates a new category model or updates the model of respective category.}}
	\label{fig:simulated_user_architecture}
\end{figure}

\noindent
{In this round of evaluation, the agent starts with no previous knowledge. The teacher randomly selects three object views of a category and \texttt{teaches} that category to the agent. The model the agent creates based on these instances, is stored in the agent's memory. After teaching each category, the teacher tests the agent to see if he has learned the new category and has not forgot the previously learned categories. This is done by \texttt{asking} the agent to recognize never-seen-before instances of all previously learned categories. In this experiment, when the agent incorrectly recognize a test instance, the teacher provides \texttt{corrective feedback}. This feedback allows the agent adjusts its category model using the mistaken instance -- this way, the agent incrementally learns new object categories using as few as possible instances.} 

The recognition performance of the agent is continuously evaluated using a sliding window over the last $3n$ {question/correction} iterations, where $n$ is the number of learned categories. We called this measure as \texttt{protocol accuracy}.
To compute the protocol accuracy, if the number of {question/correction} iterations since the last category was introduced, is less than $3n$, all results are taken into account. If the protocol accuracy of the agent exceeds a certain threshold ($\tau = 0.67$, meaning accuracy is at least twice the error rate), {the simulated teacher randomly selects a new category and teaches it to the agent.} If the agent fails to reach the protocol accuracy threshold within a certain amount of iterations (e.g., 100 iterations) since the last category was introduced, the teacher infers that the agent is not able to learn more categories at this point and stops the experiment. It is also possible that the agent learns all existing object categories and keep the protocol accuracy above the threshold. In this case, the experiment is also stopped and the stopping condition is \textit{``lack of data''}. {Since the order of teaching new categories and instances may have an effect on the performance of the agent, we have performed ten experiments for each approach. For example, if the simulated teacher introduces an apple right after a tomato, it would be harder to recognize this new object than when a banana followed the tomato are introduced. In order to compare methods fairly, the simulated teacher shuffles data at the beginning of each round of experiments and uses the same order of object categories and instances for teaching and testing all the methods. Furthermore, we re-implemented all of the selected base-lines ourselves in order to ensure a fair comparison.}

\textbf{Object Dataset and Evaluation Metrics:}
For this round of evaluations, Washington RGB-D Object dataset~\cite{lai2011large} is used. This dataset consists of $250,000$ views from 300 common household objects in 51 categories. The performance evaluation of an open-ended learning system cannot be limited to the classical evaluation metrics (accuracy, precision and recall). Therefore, we use three recently introduced metrics~\cite{oliveira2015concurrent} to evaluate the learning performance. They are including: (\emph{i}) the average number of learned categories at the end of experiments (ALC), an indicator of \emph{how much the system is able to learn in an open-ended setting}; (\emph{ii}) average number of question/correction iterations (QCI) required to learn those categories, and the average number of stored instances per category (AIC), which represent \emph{time and memory resources required to learn those categories}-- therefore, a lower value is better as it means that less instances and thus less computing power were needed to learn an object category; (\emph{iii}) Global Classification Accuracy (GCA), describes the accuracy of the agent over the whole run, and the Average Protocol Accuracy (APA), represents the average accuracy over all  windows. Therefore, GCA and APA are indicator of \emph{how well the system learns}.

\textbf{Results:} We have conducted an extensive set of open-ended experiments and comparisons to highlight the effectiveness of the proposed approach. 
In particular, we compared OrthographicNet with five recently proposed open-ended learning approaches including RACE~\cite{oliveira20163d}, BoW~\cite{kasaei2018towards} based on a $L_2$ based nearest neighbour classifier, an updated version of the online LDA~\cite{hoffman2010online} (here refer to as Open-ended LDA), Local-LDA~\cite{kasaei2016hierarchical} and GOOD~\cite{kasaei2018perceiving}. Furthermore, since our approach with \emph{MobileNet-v2}, \emph{VGG16-fc1} and \emph{ResNet50} demonstrated a good performance in previous round of evaluation, we have included all of them in this round of evaluation. 
\begin{figure}[!t]
	\centering
	\includegraphics[width=\linewidth, trim= 6cm 0cm 5cm 0cm,clip=true]{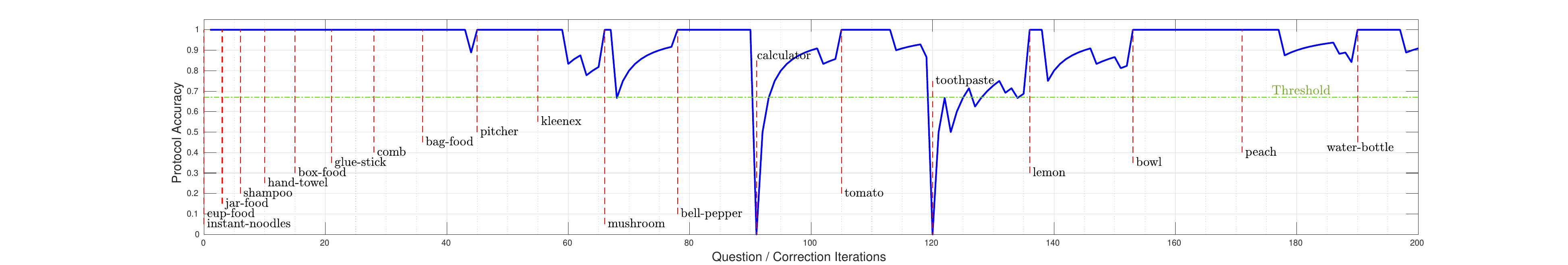}\\
	\includegraphics[width=\linewidth, trim= 6cm 0cm 5cm 0cm,clip=true]{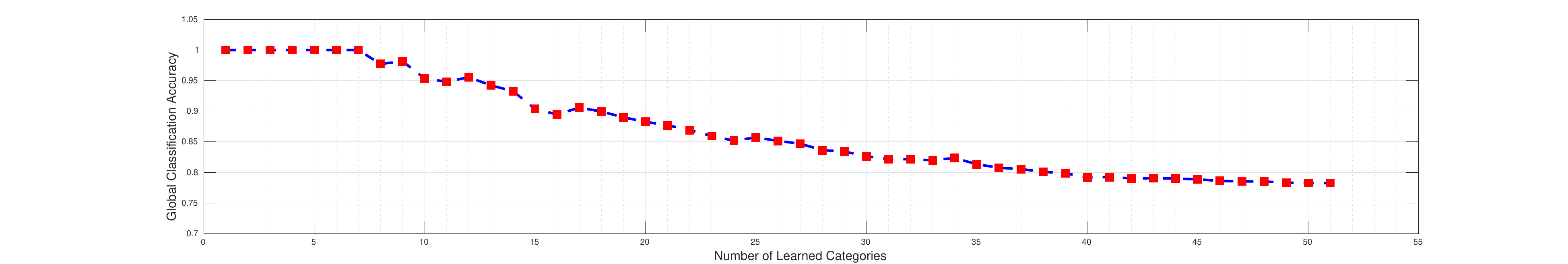}\\
	\includegraphics[width=\linewidth, trim= 6cm 0cm 5cm 0cm,clip=true]{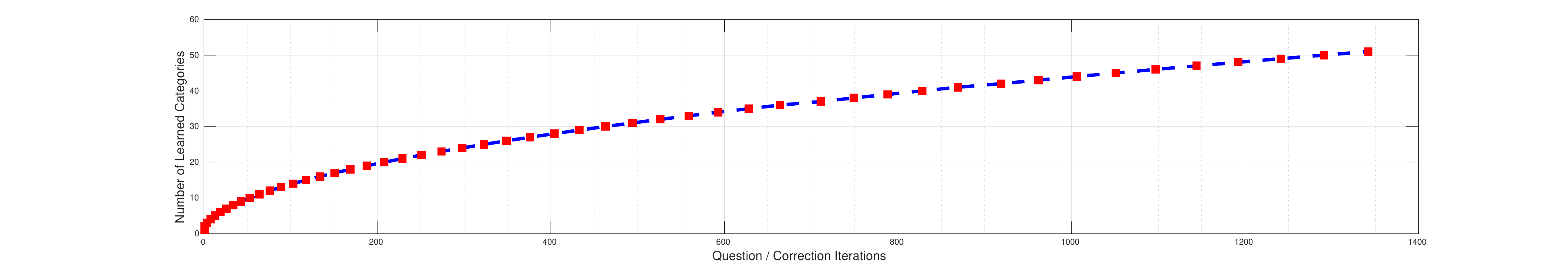}\\
	\includegraphics[width=\linewidth, trim= 6cm 0cm 5cm 0cm,clip=true]{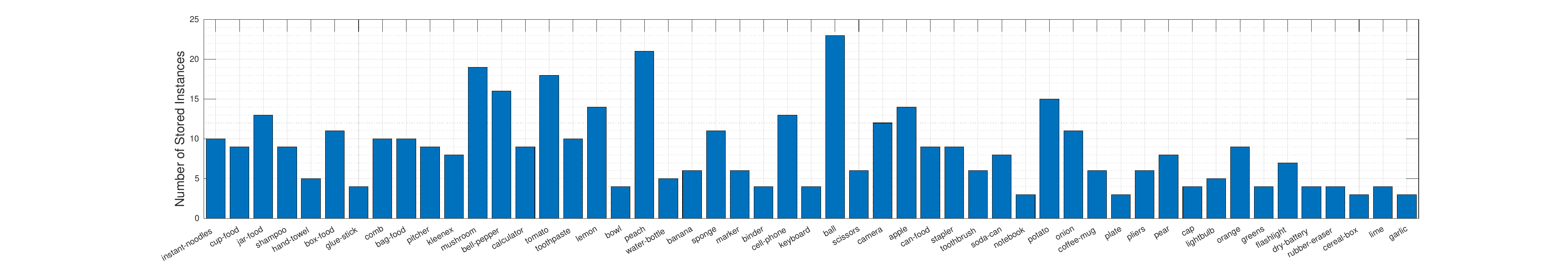}\\
	
	\caption{Summary of the first open-ended experiment of \emph{OrthographicNet-MobileNetV2}: (\emph{top}) Evolution of protocol accuracy over first 200 iterations. In this experiment, the protocol threshold is set to $0.67$, as shown by the green horizontal dashed line. Once the agent has classified all learned categories at least once, and the protocol accuracy is higher than the threshold, a new category is introduced, which is shown by a red dashed line and a category name. (\emph{second}) This plot shows global classification accuracy as a function of number of learned categories for the same experiment. It can be seen that the agent was able to stay above the protocol threshold during the entire experiment, indicating the agent can learn many more categories. (\emph{third}) This graph represents how fast does the agent learn by representing the number of learned categories as a function of the number of question/correction iterations. (\emph{bottom}) This graph shows the number of instances stored in each category, i.e., the three instances provided at the introduction of the category together with the instances that had to be corrected somewhere along the experiment run. The ball category apparently was the most difficult one, requiring the largest number of instances. }
	\label{open_ended_learning}
\end{figure}

\begin{figure}[!b]
	\begin{tabular}[width=\textwidth]{cc}
		\hspace{-5mm}
		\includegraphics[width=0.5\linewidth, trim= 0.2cm 0cm 0.2cm 0cm,clip=true]{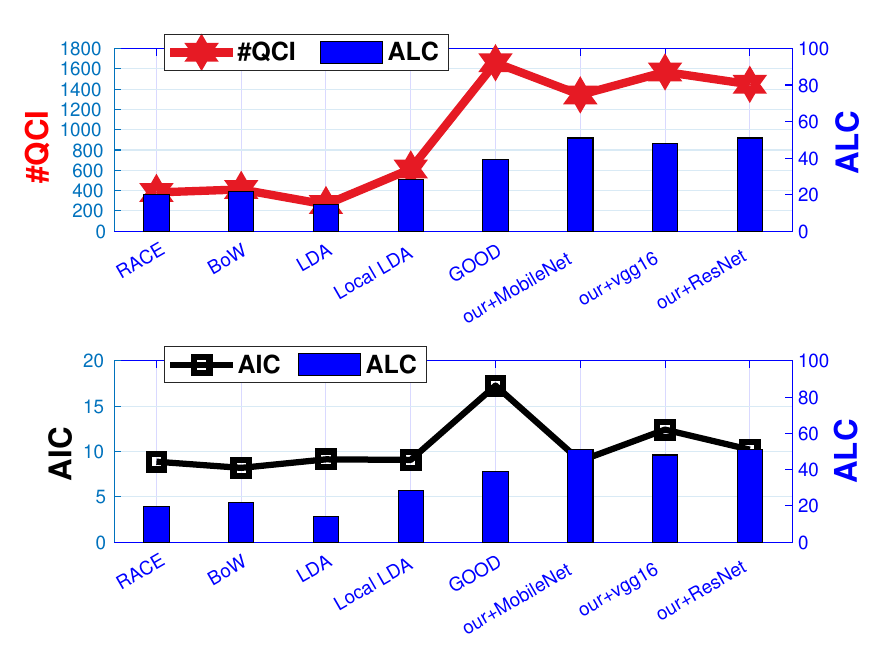} & \hspace{-4mm}
		\includegraphics[width=0.5\linewidth, trim= 0.5cm 0.0cm 0.2cm 0cm,clip=true]{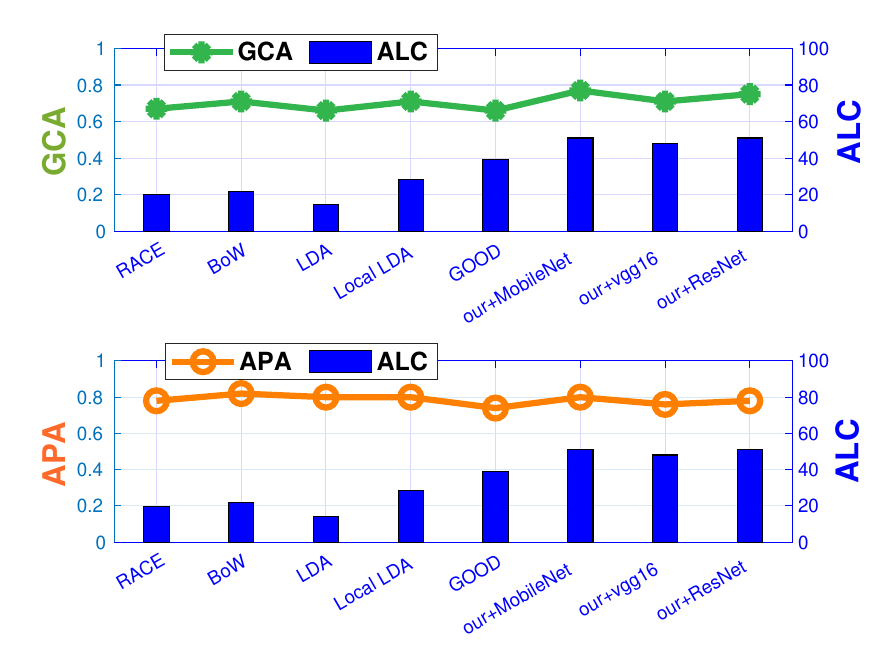}
	\end{tabular}
	\caption{Summary of open-ended evaluations: (\textit{left-top})
		shows the number of question/correction iterations (QCI) required to learn
		a certain number of categories; (\textit{left-bottom}): Average number of stored
		instances per category (AIC) and average number of learned categories
		(ALC) at the end of an experiment as an indicator of how much memory
		does each approach take?; The right-top and right-bottom plots correlate the
		global classification accuracy (GCA), and average protocol accuracy (APA)
		with the ALC as an indicator of how well does it learn?.}
	\label{fig:open_ended_evaluations}       % Give a unique label
\end{figure}

To better understand the idea, the performance of the \emph{OrthographicNet-MobileNetV2} in the initial 200 iterations of the first open-ended experiment is plotted in Fig.~\ref{open_ended_learning} (\emph{top}). In this experiment, the simulated user progressively estimates the recognition accuracy of the system and, in case this accuracy exceeds a given threshold (marked by the horizontal green line), introduces an additional object category (marked by the vertical red lines and labels). The simulated teacher also provides corrective feedback in case of misclassification. This way, the system is trained and at the same time the accuracy of the system is continuously estimated. This plot clearly represents the different steps of
the simulated teaching process. In this experiment, categories that are very similar in term of shape is harder to learn, think for example about rounded objects (lime, bell pepper, onion, tomato, apple, etc.). Fig.~\ref{open_ended_learning} (\emph{second}) shows the global classification accuracy (i.e. the accuracy since the beginning of the experiment) as a function of the number of learned categories. In this figure, we can see that the global classification accuracy decreases as more categories are learned. This is expected since the number of categories known by the system makes the classification task more difficult. Fig.~\ref{open_ended_learning} (\emph{third}) shows the number of learned categories as a function of the protocol iterations. This gives a measure of how fast the learning occurred in each of the experiments. This experiment concluded prematurely due to the ``\emph{lack of data}'' (i.e., no more categories available in the dataset), indicating the potential for learning many more categories. Fig.~\ref{open_ended_learning} (\emph{bottom}) represents the number of instances stored in the models of all of the categories of the mentioned experiment. These are the three instances provided at the introduction of the category together with any instances that had to be corrected somewhere along the experiment run. Ball category apparently was the most difficult category, requiring the largest number of instances. However do note that instances that were introduced near the end of the experiment have been tested less, which is clearly visible in a general trend for less instances to be included for categories appearing later.
\begin{table}[!t]
	\begin{center}
		\caption {Result of open-ended evaluations using RGB-D dataset~\cite{lai2011large}.}
		\resizebox{0.95\linewidth}{!}{
			\begin{tabular}{ |c|c|c|c|c|c| }
				\hline
				Approaches & \#$\operatorname{QCI}$ & $\operatorname{ALC}$ & $\operatorname{AIC}$ & $\operatorname{GCA}$ & $\operatorname{APA}$ \\
				\hline \hline
				RACE~\cite{oliveira20163d} &382.10 &19.90 & 8.88 & 0.67 & 0.78\\
				\hline 
				BoW~\cite{kasaei2018towards} & 411.80 &21.80 & \textbf{8.20} & 0.71 & \textbf{0.82}\\
				\hline 
				Open-Ended LDA~\cite{hoffman2010online} & \textbf{262.60} &14.40 & 9.14& 0.66& 0.80\\
				\hline
				Local-LDA~\cite{kasaei2016hierarchical} & 613.00 &28.50& 9.08& 0.71 &0.80\\
				\hline
				GOOD~\cite{kasaei2018perceiving}& 1659.20 & 39.20 & 17.28 & 0.66  & 0.74 \\
				\hline \hline 
				ours-\emph{MobileNet-v2}\cblue{$^{\textbf{(*)}}$} & 1342.60 & \textbf{51.00} & 8.97 & \textbf{0.77} & 0.80\\
				\hline 
				ours-\emph{VGG16-fc1}\cred{$^{\textbf{(\#)}}$} & 1568.70	& 48.10 & 12.42 & 0.71 & 0.76\\
				\hline
				ours-\emph{ResNet50}\cblue{$^{\textbf{(*)}}$} & 1446.60 & \textbf{51.00} & 10.21 & 0.75 & 0.78
				\\
				\hline

		\end{tabular}}
		\label{table_open_ended_evaluations}
		\\\scriptsize{\cblue{$^{\textbf{(*)}}$} Stopping condition was ``\emph{lack of data}''. \cred{$^{\textbf{(\#)}}$} In 6 out of 10 experiments, stopping condition was ``\emph{lack of data}''}
	\end{center}
\end{table}

A detailed summary of all experiment is reported in the Table~\ref{table_open_ended_evaluations} and Fig.~\ref{fig:open_ended_evaluations}. One important observation is that the scalability of OrthographicNet was much better than the other approaches. In particular, our approach with \emph{MobileNet} and \emph{ResNet50} learned all 51 categories in all 10 experiments. All of those experiments concluded prematurely due to the ``\emph{lack of data}'', i.e., no more categories available in the dataset, indicating the potential for learning many more categories. OrthographicNet with \emph{VGG16-fc1} also obtained acceptable scalability (i.e., in six out of 10 experiments, the agent could learn all categories) while the scalability of other approaches was much lower than ours. Our approach with \emph{MobileNet}, on average, learned all the categories faster than the other CNN approaches. QCI measures \emph{how fast} the learning occurred in each of the experiments. It was observed that the longest experiments were \emph{OrtographicNet-VGG16-fc1}, and the shortest ones were Open-Ended LDA. In particular, on average, the agent with Open-Ended LDA was able to learn 14.40 categories in 262.60 iterations, while the agent with \emph{OrtographicNet-VGG16-fc1} continued for 1568.70  iterations to learn 48.10 categories. By comparing all approaches, it is clear that RACE, BoW, and \emph{OrthographicNet-MobileNet-v2}, on average, stored less than nine instances per category. Although RACE and BoW stored fewer instances per category (AIC) than \emph{OrthographicNet-MobileNet-v2}, the difference is minor (less than one instance per category). The discriminative power of OrthographicNet is much better than RACE and BoW. Specifically, our approach learned 31.10 and 29.20 categories more than RACE and BoW approaches, respectively. The right column of Fig.~\ref{fig:open_ended_evaluations} correlates the global classification accuracy (GCA, \emph{top-right}) and average protocol accuracy (APA, \emph{bottom-right}), obtained by the evaluated approaches, with the average number of learned categories (ALC). Our approach with \emph{MobileNet-v2} achieved the best GCA with stable performance. BoW achieved better performance than our approaches regarding APA. This is expected since BoW learned fewer categories, and it is easier to get better APA in fewer categories. In particular, our approach with \emph{MobileNet-v2} was able to learn all 51 categories, on average, while the other approaches learned less than 40 categories. It can be concluded that  \emph{OrthographicNet-ModelNet-v2} achieved the best performance.

{The \texttt{protocol\_threshold} parameter, $\tau$, defines how good the agent should learn object categories. For example, $\tau=0.67$ means the recognition accuracy is at least twice better than the error. To explore the impact of protocol threshold, an additional set of experiments with different values of $\tau \in [0.7, 0.8, 0.9]$ was evaluated using the best system configuration (OrthographicNet+MobileNet-v2). It should be noted that the order of introducing categories and instances is the same for all experiments. Refer to Table~\ref{table_open_ended_evaluations_diffrent_tau} for the impact that changing $\tau$ has on system outcomes. We can see that, as $\tau$ increases, the GCA and APA increases as
well. However, this comes at the cost of the amount of learned categories. The fact that the orthographicNet with $\tau = 0.9$ yields better accuracies, can be explained by Fig.~\ref{open_ended_learning} (second plot from top), which clearly shows that the amount of learned categories has a negative correlation with global classification accuracy.}

\begin{table}[t]
	\begin{center}
		\caption {{Result of open-ended evaluations with different values for the protocol threshold~\cite{lai2011large}.}}
		\resizebox{0.75\linewidth}{!}{
			\begin{tabular}{ |c|c|c|c|c|c| }
				\hline
				$\tau$ & \#$\operatorname{QCI}$ & $\operatorname{ALC}$ & $\operatorname{AIC}$ & $\operatorname{GCA}$ & $\operatorname{APA}$ \\
				\hline \hline
				0.70 & \textbf{1384} & \textbf{51} & \textbf{9.90}  & 0.77 & 0.80\\
				\hline 
				0.80 & 1943 & \textbf{51} & 10.41 & 0.81 & 0.84\\
				\hline
				0.90 & 1972 & 34 & 10.82 & \textbf{0.87} & \textbf{0.93}
				\\
				\hline
		\end{tabular}}
		\label{table_open_ended_evaluations_diffrent_tau}
	\end{center}
\end{table}

%%%%%%%%%%%%%%%%%%%%%%%%%%%%%%%%%%%%%%%%%%%%%%%%%%%%%
\subsection{Evaluation of Real-Time Performance}
%%%%%%%%%%%%%%%%%%%%%%%%%%%%%%%%%%%%%%%%%%%%%%%%%%%%%
We carried out three real-word demonstrations to show the real-time performance of the proposed approach. {For these experiments, we integrated our work into the robotic system presented in~\cite{oliveira20163d}, and used Kinect V1 sensor to perceive the environment}.

\begin{figure}[!b]
	\centering
	\includegraphics[width=0.95\columnwidth, trim= 0cm 0cm 0.0cm 0cm,clip=true]{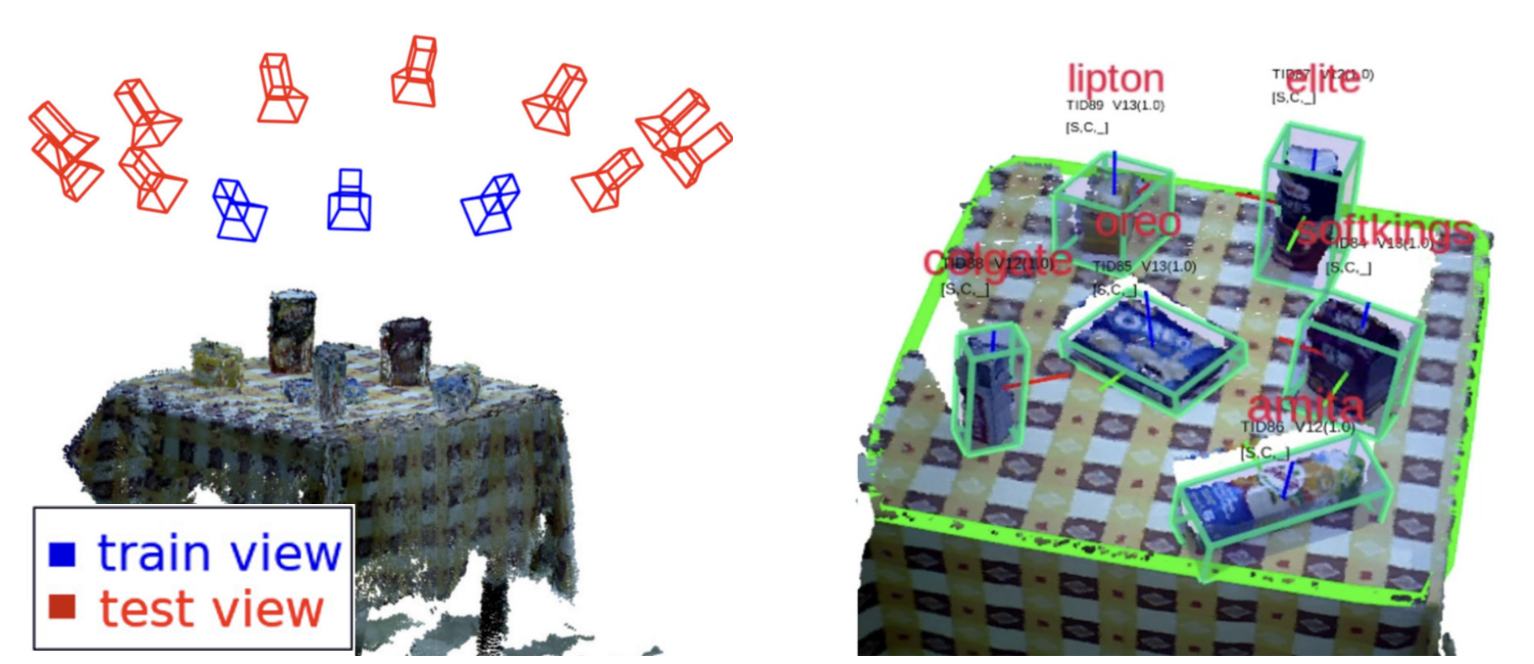}
	\caption{Domestic environment demonstration: (\textit{left}) Experimental setup: we use three consecutive scenes for training (blue cameras) and use the remaining 10 scenes for testing (red cameras). (\textit{right}) An example of object recognition results on the first test scene.}
	\label{real_demo}
\end{figure}

%%%%%%%%%%%%%%%%%%%%%%%%%%%%%%%%%%%%%%%%%%%%%%%%%%%%%
\textbf{Domestic Environment:}
 We have used the Domestic Environment Dataset~\cite{doumanoglou2016recovering} for this round of evaluation. This dataset consists of several 3D scenes containing a table and a set of table-top objects including \emph{amita}, \emph{colgate}, \emph{lipton}, \emph{elite}, \emph{oreo} and \emph{softkings}. As shown in Fig.~\ref{real_demo}, each scene has been captured under different levels of clutter and occlusion and contains a various set of objects in different position and orientation. Therefore, it is a suitable dataset to evaluate the robustness of the proposed approach to clutter and occlusion, as well as its scale- and pose-invariant properties. 

In this evaluation, the robot begins without any prior knowledge; therefore, it recognizes all detected objects as \textit{unknown}. A human-user teaches all object categories to the robot using the provided human-robot interaction interface. This way, the robot conceptualizes all object categories in an online fashion using the representation of extracted training objects' views (see Fig.~\ref{real_demo}). Afterwards, we test the robot by the remaining ten scenes (shown by red cameras in Fig.~\ref{real_demo} (\textit{left}). It was observed that the robot could recognize all objects correctly based on the learned knowledge from the three training scenes (Fig.~\ref{real_demo} \textit{right}). It should be mentioned that some misclassification also happened during the evaluation. At some points, the object tracking module did not track the point cloud of the object precisely. Therefore, different portions of the target object were excluded, and misclassification happened as a result. 

\begin{figure*}[!t] 
	\center
	\includegraphics[width=\linewidth, trim= 0cm 0cm 0.0cm 0cm,clip=true]{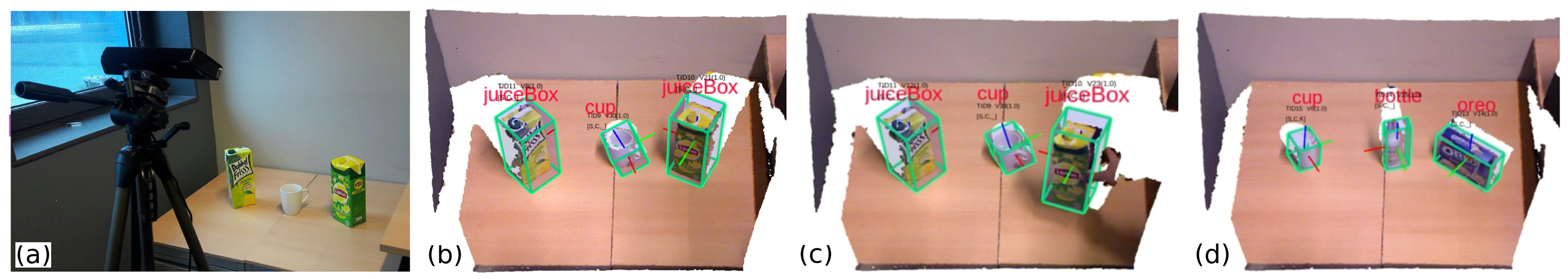}
	\caption{ \emph{Serve\_a\_drink} scenario: (\emph{a}) experimental setup; (\emph{b}) This snapshot shows the proposed approach supports both batch and open-ended learning cases; The system gained knowledge about \emph{juiceBox} using batch learning and learns about \emph{cup} category in an open-ended manner; (\emph{c}) A user then moves the \emph{juiceBox} and the system can track and recognize it correctly; (\emph{d}) The user removes all objects from the table and adds three new objects including \emph{oreo}, \emph{bottle}, and a new \emph{cup}. The system had prior knowledge about \emph{oreo} and learned about \emph{bottle} category in an online manner. The system recognizes all object correctly.}
	\label{fig:real_demo_serve_a_drink}
\end{figure*}

We later tested the robot in two new contexts. The first context contains six instances  \emph{lipton}, \emph{oreo}, and \emph{amita} categories.  It was seen that the robot recognized all instances by using the learned knowledge from the previous context. In this context, we mainly evaluated the scale- and pose-invariant properties of the \emph{OrthographicNet}.  The robot recognized both \emph{amita} instances correctly. In the second context, all scenes contain four instances of two (\emph{lipton} and \emph{softkings}) categories, where the two instances of lipton occluded the softking objects. Besides, instances of these categories have a very similar shape, and therefore, make the object recognition task harder for the robot. We observed that the robot correctly recognized all instances using the learned knowledge from the initial context (i.e., three consecutive scenes). This evaluation shows the robustness and descriptive power of the proposed approach. A video of this evaluation is accessible online at \href{https://youtu.be/JEUb-Q7TbJQ}{\small \cblue{https://youtu.be/JEUb-Q7TbJQ}}

%%%%%%%%%%%%%%%%%%%%%%%%%%%%%%%%%%%%%%%%%%%%%%%%%%%%%%%%%%%%%%%%%%%%%%%%%%%%%%%%%%%%%%%%%%%%%%%
\textbf{``Serve\_A\_Drink'' scenario:}
In this scenario, there is a table in front of a Kinect sensor as shown in Fig.~\ref{fig:real_demo_serve_a_drink} (\emph{a}). For object categories including \emph{juiceBox}, \emph{oreo}, \emph{cup} and \emph{bottle} are used in this round of evaluation. In contrast with the previous demonstration, the robot initially knows \emph{juiceBox} and \emph{oreo} categories, which have been learned from batch data, and does not have any knowledge about the \emph{bottle} and \emph{cup} categories.  Throughout this evaluation, a human-user puts objects on top of the table and teaches new categories to the robot. The robot should demonstrate the capability of recognizing instances of known categories, and also the capability of learning about new object categories in a supervised and open-ended manner. Figure~\ref{fig:real_demo_serve_a_drink} and the following description explains the behavior of the proposed approach:

\begin {itemize}
\item The user presents a \emph{cup} object to the system and provides the respective category labels (TrackID9 as a \emph{cup}). The system conceptualizes the \emph{cup} category and TrackID9 is correctly recognized. 
\item The instructor places a \emph{juiceBox} on the table. The system has learned about \emph{juiceBox}  category from batch data, therefore, TrackID10 is recognized properly. An additional \emph{juiceBox} is placed at the left-side of the table. Tracking is initialized and the \emph{juiceBox} is recognized accurately (Fig.~\ref{fig:real_demo_serve_a_drink} \emph{b}).
\item The instructor moves the right \emph{juiceBox} for a while to show the real-time performance of the \emph{OrthographicNet} for performing object recognition and pose estimation concurrently (Fig.~\ref{fig:real_demo_serve_a_drink} \emph{c}).

\item Later, the user removes all objects from the scene; no objects are visible;
an \emph{oreo} and a \emph{bottle} enter the scene. They are detected and assigned to TrackID13 and TrackID14 respectably. Because there is no prior knowledge about \emph{bottle} category, a misclassification happened. TrackID14 is labelled as a \emph{bottle}; the system first conceptualizes the \emph{bottle} category and then recognizes it correctly.

\item  Another \emph{cup} is placed on the table. This particular \emph{cup} had not been seen before, but it recognizes correctly since the system learned about \emph{cup} category earlier (Fig.~\ref{fig:real_demo_serve_a_drink} \emph{d}).
\end{itemize}

\noindent
This demonstration shows that apart from batch learning, the robot can also learn about new object categories in an open-ended fashion. Furthermore. it shows that the proposed approach is capable of recognizing objects in various positions. A video of this demonstration is available online at: \href{https://youtu.be/sFJFc_lnzHQ}{\small \cblue{https://youtu.be/sFJFc\_lnzHQ}}

\begin{figure} [!b]
    \centering
	\includegraphics[width=\linewidth, trim= 0cm 0cm 0.0cm 0cm,clip=true]{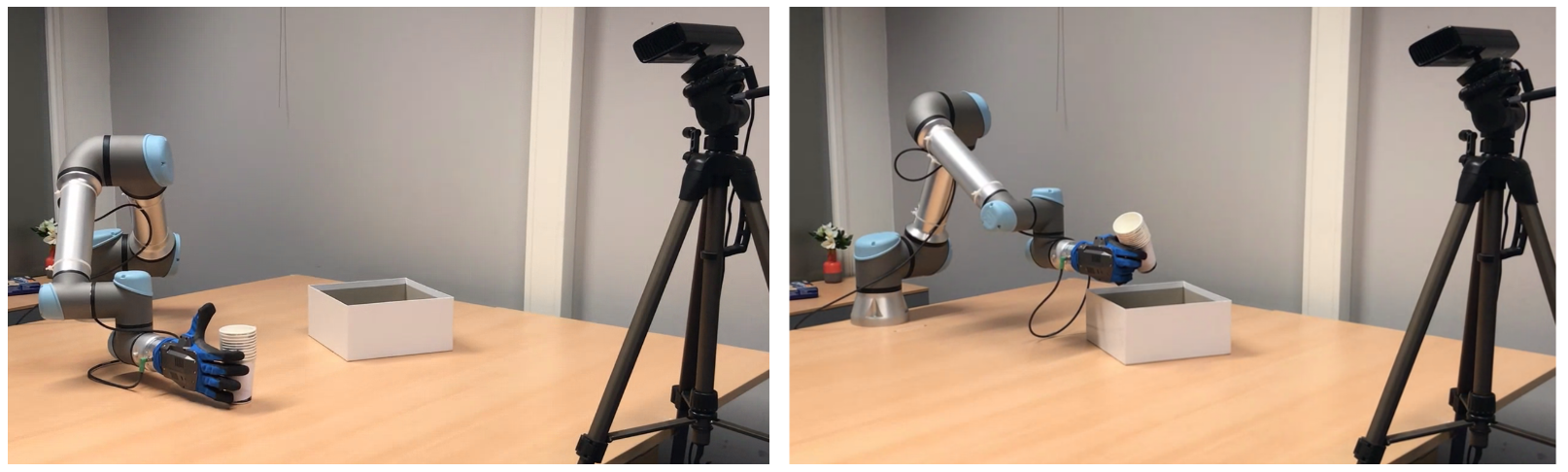}
    \caption{{Our experimental setup for the real-robot experiment: it consists of a Kinect camera, a UR5e robotic-arm and a Qbhand as the primary sensory-motor embodiment for perceiving and acting upon its environment.}}
    \label{setup_orthographic}
\end{figure}

\begin{figure*}[!t] 
	\center
	\includegraphics[width=.95\linewidth, trim= 0cm 0cm 0.0cm 0cm,clip=true]{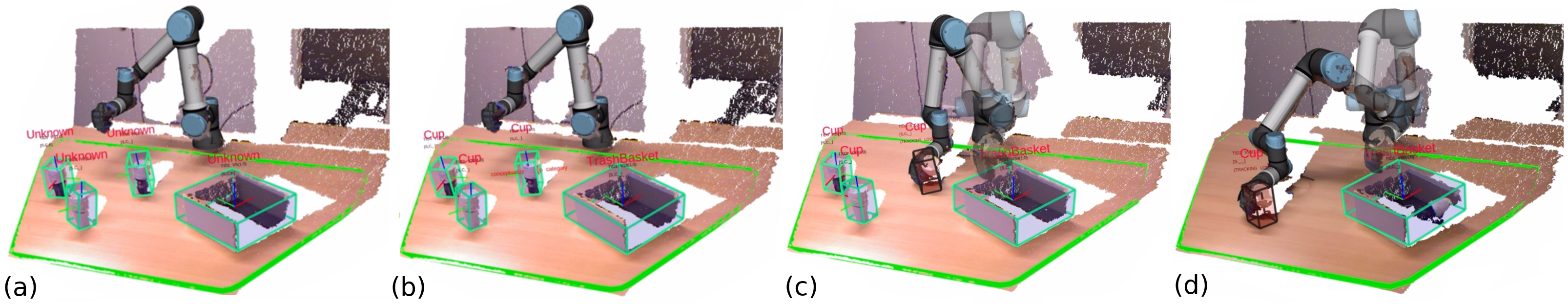}
	\caption{The \textit{clear\_table} scenario; (\textit{a}) All table-top objects are detected and highlighted by green bounding boxes. The local reference frame of each object represents the pose of the object. The object reference frame is initially estimated by the proposed approach and then tracked by an object tracking module. In this demonstration, the robot has no prior knowledge about any object. Therefore, all objects are recognized as \textit{unknown}, shown by red on top of each object.	(\textit{b}) A user then teaches the \textit{Cup} and \textit{TrashBasket} categories to the robot, and all objects are then correctly recognized. Afterwards, the user instruct the robot to perform clear\_table task. Towards this goal, in each execution cycle, the robot should find out the nearest \textit{Cup} and go to its pre-grasp area (\textit{c}); It then grasps the \textit{Cup} and moves it on top of the \textit{TrashBasket} and releases it (\textit{d}). This procedure is repeated several times to remove all \textit{Cups} from the table.}
	\label{fig:real_demo_clear_table}
	%\vspace{-2mm}
\end{figure*}

\textbf{Robotic application:} {We performed a real-robot experiment in the context of clear-table task. Our experimental setup is shown in Fig.~\ref{setup_orthographic}. In this experiment, the robot should detect the pose and recognize all objects including \textit{Cups} and the \textit{TrashBasket}. Then, the robot should clean all Cups from the table by putting them into the basket.} At the beginning of this experiment, the set of categories known to the system is empty and the system recognizes all table-top objects as \textit{Unknown} (Fig.~\ref{fig:real_demo_clear_table} \textit{a}). A user interacts with the system by teaching \textit{Cup} and \textit{TrashBasket} categories. The system conceptualizes both categories and later recognizes them properly (Fig.~\ref{fig:real_demo_clear_table} \textit{b}). This illustrates the process of acquiring categories in an open-ended fashion with user mediation. Later, the user instructs the robot to perform a \textit{clear\_table} task. In this experiment, the robot must be able to detect the pose and recognize all active objects. Afterwards, in each execution cycle, it has to grasp the nearest \textit{Cup} and then transport it on top of the TrashBasket and release it. This sequence is shown in Fig.~\ref{fig:real_demo_clear_table} (\textit{c}) -- (\textit{d}). A video of this session is available
online at: \href{https://youtu.be/4n2v8LmB9D0o}{\small{\cblue{https://youtu.be/4n2v8LmB9D0}}}

%%%%%%%%%%%%%%%%%%%%%%%%%%%%%%%%%%%%%%%%%%%%%%%%%%%%%%%%%%%%%%%%%%%%%%%%%%%%%%%%%%%%%%
\section{Conclusion} 
\label{sec:conclusion}
In this paper, we propose a deep transfer learning based approach for 3D object recognition in open-ended domains named \emph{OrthographicNet}. This approach provides a good trade-off between descriptiveness, computation time and memory usage, allowing concurrent object recognition and pose estimation. \emph{OrthographicNet} computes a unique and repeatable global object reference frame and three scale-invariant orthographic projections for a given object. The orthographic projects are then fed as input to three CNNs to obtain a view-wise deep feature. The obtained features are then merged using an element-wise max pooling layer to form a  rotation-invariant global feature for the given object. 
A set of experiments were carried out to assess the performance of \emph{OrthographicNet} and compare it with other state-of-art with respect to several characteristics including descriptiveness, scalability and memory usage. We have shown that \emph{OrthographicNet} can achieve performance better than the selected state-of-the-art and is suited for real-time robotic applications. In the continuation of this work, we would like to consider color information~\cite{kasaei2020investigating,keunecke2020combining} and use orthographic projections for simultaneous object recognizing and grasp affordance detection.

\bibliographystyle{IEEEtran}
\bibliography{references}

\begin{IEEEbiography}[{\includegraphics[width=1in,height=1.25in,clip,keepaspectratio]{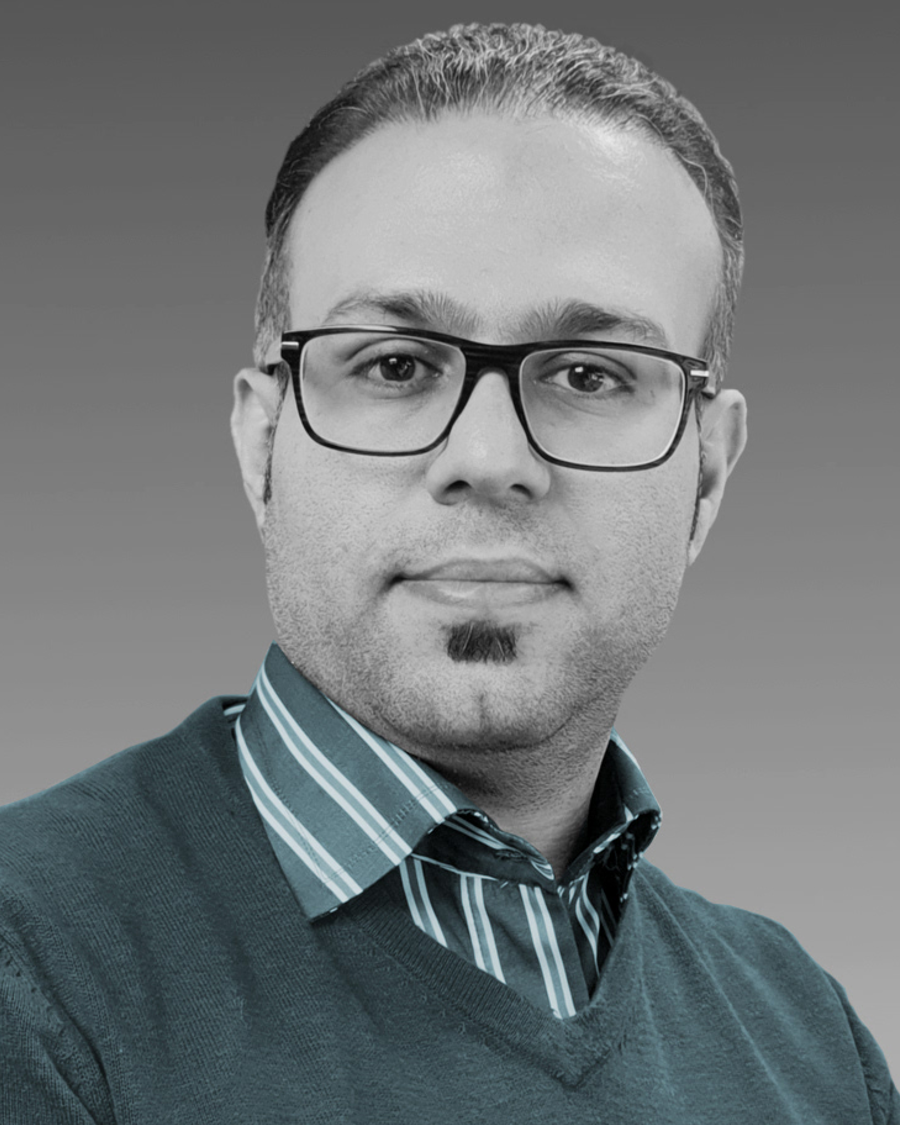}}]{Hamidreza Kasaei} is an Assistant Professor in the Department of Artificial Intelligence at the University of Groningen, the Netherlands. Hamidreza received a Ph.D. degree in Robotics from the University of Porto (MAP-i), Portugal, in 2018. He was a visiting researcher at Imperial College London, UK, in 2016. He has been serving as Associate Editor for IEEE International Conference on Robotics and Automation (ICRA) since 2019. Kasaei's research interests focus on the intersection of robotics, machine learning, and machine vision. His main research goal is to achieve a breakthrough by enabling robots to incrementally learn from past experiences and safely interact with human users using open-ended machine learning techniques.
\end{IEEEbiography}

\end{document}